%% file: main.tex
\documentclass[10pt,twocolumn,letterpaper]{article}

\usepackage[pagenumbers]{cvpr} 
\usepackage{makecell}

\input{preamble}
\definecolor{cvprblue}{rgb}{0.21,0.49,0.74}
\usepackage[pagebackref,breaklinks,colorlinks,allcolors=cvprblue]{hyperref}

\def\paperID{13} 
\def\confName{CVPR}
\def\confYear{2026}

\title{BlazeFL: Fast and Deterministic Federated Learning Simulation}

\author{Kitsuya Azuma\\
Institute of Science Tokyo\\
Tokyo, Japan
\and
Takayuki Nishio\\
Institute of Science Tokyo\\
Tokyo, Japan
}

\begin{document}
\maketitle
\renewcommand\thefootnote{}
\footnotetext{This paper has been accepted to the FedVision at CVPR 2026 (CVPRW). $\copyright$ 2026 IEEE. This is the author's accepted version of the paper.}
\renewcommand\thefootnote{\arabic{footnote}}
\input{sec/0_abstract}    
\input{sec/1_introduction}
\input{sec/2_background_and_related_work}
\input{sec/3_system_design}
\input{sec/4_evaluation}
\input{sec/5_limitations}
\input{sec/6_conclusion}
{
    \small
    \bibliographystyle{ieeenat_fullname}
    \bibliography{main}
}


\end{document}

%% file: sec/0_abstract.tex
\begin{abstract}
Federated learning (FL) research increasingly relies on single-node simulations with hundreds or thousands of virtual clients, making both efficiency and reproducibility essential. Yet parallel client training often introduces nondeterminism through shared random state and scheduling variability, forcing researchers to trade throughput for reproducibility or to implement custom control logic within complex frameworks. We present BlazeFL, a lightweight framework for single-node FL simulation that alleviates this trade-off through free-threaded shared-memory execution and deterministic randomness management. BlazeFL uses thread-based parallelism with in-memory parameter exchange between the server and clients, avoiding serialization and inter-process communication overhead. To support deterministic execution, BlazeFL assigns isolated random number generator (RNG) streams to clients. Under a fixed software/hardware stack, and when stochastic operators consume BlazeFL-managed generators, this design yields bitwise-identical results across repeated high-concurrency runs in both thread-based and process-based modes. In CIFAR-10 image-classification experiments, BlazeFL substantially reduces execution time relative to a widely used open-source baseline, achieving up to 3.1$\times$ speedup on communication-dominated workloads while preserving a lightweight dependency footprint. 
Our open-source implementation is available at: \url{https://github.com/kitsuyaazuma/blazefl}.
\end{abstract}

%% file: sec/1_introduction.tex
\section{Introduction}
\label{sec:introduction}

\begin{figure}[t]
    \centering
    \includegraphics[width=\linewidth]{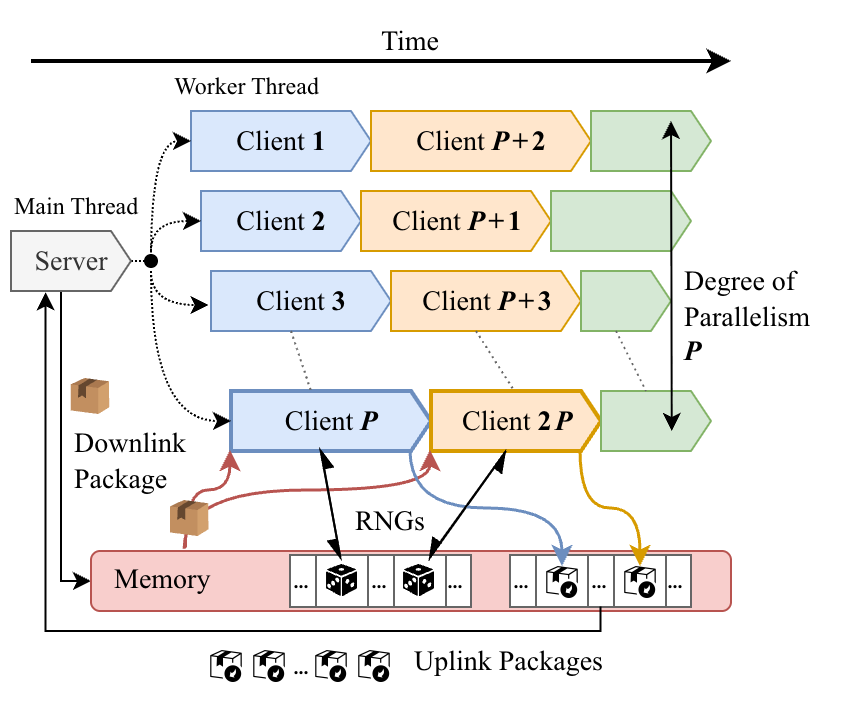}
    \caption{Architecture overview of BlazeFL. A main thread coordinates client scheduling, while worker threads execute within a shared address space, enabling server-to-client parameter broadcast and client-to-server uploads without cross-process serialization or IPC. Each client is associated with an isolated RNG stream to support deterministic repeated execution under controlled settings.}
    \label{fig:architecture}
\end{figure}

Federated learning (FL) enables model training across distributed devices without centralizing privacy-sensitive data. In practice, much FL research is first conducted through single-node simulation, where hundreds or thousands of virtual clients are repeatedly sampled, trained, and aggregated before real-world deployment. 

Particularly in computer vision tasks, which inherently involve large-scale model parameters (e.g., deep ResNets, Vision Transformers) and compute-intensive data augmentation pipelines, the overhead of inter-process communication and repeated parameter serialization severely limits simulation scalability. As datasets and model architectures grow, the runtime of these simulations becomes a major bottleneck for algorithm prototyping, ablation studies, and hyperparameter search.

To accelerate such workloads, existing FL frameworks rely on parallel execution. In conventional Python environments, however, the Global Interpreter Lock (GIL) limits true CPU parallelism for threads. Consequently, many FL systems adopt multiprocessing or distributed runtimes such as Ray~\cite{ray}. Although these designs reduce the cost of straightforward data transfer, they still introduce nontrivial overheads associated with process isolation, metadata management, and repeated parameter exchange across communication rounds.

Reproducibility presents a second challenge, particularly the ability to achieve bitwise-identical results across runs given the same seed. FL simulations contain multiple sources of stochasticity, including client sampling, data partitioning, mini-batch ordering, data augmentation, and regularization. Under parallel execution, nondeterminism can arise from shared or poorly controlled random states, as well as from completion-order-dependent aggregation, where the order of floating-point accumulation varies across runs. Even when using a well-established and carefully engineered FL framework such as Flower~\cite{flower} and FedML~\cite{fedml} with a fixed global seed, repeated runs can yield differences not only in the resulting model weights but also in the final performance of the trained model. This lack of reproducibility hinders researchers' ability to perform fine-grained analysis of the internal behavior of machine learning models across training trajectories.

These observations motivate a practical systems question: can a single-node FL simulator improve throughput while retaining controlled, repeatable execution? We present \textbf{BlazeFL}, a lightweight framework for single-node FL simulation built on Python's free-threading architecture (PEP 703~\cite{pep703}, PEP 779~\cite{pep779}). 
An overview of the framework is shown in \cref{fig:architecture}.
BlazeFL executes clients as worker threads within a single process, allowing model parameters to be exchanged through shared memory rather than cross-process serialization. BlazeFL also assigns isolated random number generator (RNG) streams to clients, reducing interference between concurrent workers. Under a fixed software/hardware stack, and when stochastic operators consume BlazeFL-managed generators, BlazeFL yields bitwise-identical results across repeated high-concurrency runs in both free-threaded and process-based modes.

The main contributions of this work are as follows:
\begin{itemize}
    \item \textbf{Shared-memory FL simulation via free-threading:} BlazeFL uses thread-based parallelism within a single process to reduce serialization and IPC overhead in single-node FL workloads.
    \item \textbf{Controlled deterministic execution:} BlazeFL assigns isolated RNG streams to clients and supports bitwise-identical repeated execution under controlled settings, as verified in our high-concurrency experiments.
    \item \textbf{Lightweight open-source design:} BlazeFL provides a minimal-dependency implementation that integrates easily with existing PyTorch-based FL pipelines and supports practical benchmarking for FL systems research.
\end{itemize}


%% file: sec/2_background_and_related_work.tex
\section{Background and Related Work}
\label{sec:background_and_related_work}

\subsection{Parallel Execution and Communication Overhead in FL Simulation}

Large-scale FL simulation repeatedly executes many clients and exchanges model parameters across communication rounds. In conventional Python environments, the Global Interpreter Lock (GIL) limits true CPU parallelism for threads. As a result, many existing FL systems adopt multiprocessing or external distributed runtimes. Frameworks such as Flower~\cite{flower}, FedML~\cite{fedml}, and pfl-research~\cite{pfl-research} build on backends including Ray~\cite{ray}, MPI~\cite{mpi}, NCCL~\cite{nccl}, and Horovod~\cite{horovod}. These runtimes can improve scalability and flexibility, but for single-node simulation they also introduce process boundaries, runtime orchestration, and additional dependencies that may increase overhead in communication-intensive workloads.

A separate line of practice is to reduce process-to-process transfer costs through shared memory, for example by placing tensors in shared memory with PyTorch~\cite{pytorch}. Such approaches can reduce parameter serialization overhead, but they still require explicit process management and careful coordination of shared state across workers.

BlazeFL targets a narrower setting: single-node FL simulation, which remains the primary environment for rapid algorithm prototyping, hyperparameter search, and ablation studies prior to real-world deployment. Rather than using multiple processes as the default execution model, BlazeFL leverages Python's free-threading support (PEP 703~\cite{pep703}, PEP 779~\cite{pep779}) to execute clients within a single process. This design allows server-to-client parameter broadcast and client-to-server uploads to occur through shared memory, reducing cross-process serialization and IPC overhead. The goal is not to replace general distributed FL runtimes, but to provide a lightweight execution path for communication-intensive single-node experiments.

\subsection{Reproducibility under Parallel Execution}

Reproducibility in FL simulation is challenging because randomness enters at multiple stages, including client sampling, data partitioning, mini-batch ordering, data augmentation, and stochastic regularization. Under parallel execution, nondeterminism may arise for at least two reasons. First, workers may share, duplicate, or inconsistently restore random states, causing the mapping between random-number consumption and client execution to depend on scheduling. Second, even when seeds are fixed, completion-order-dependent aggregation can introduce round-to-round differences because floating-point addition is not associative.

Accordingly, simply setting a global seed once is often insufficient for reproducible parallel simulation. More robust approaches require explicit management of per-worker random states across communication rounds or the use of isolated per-client generators.

BlazeFL adopts the latter strategy by associating each client with a dedicated RNG stream, thereby decoupling client-local stochasticity from worker scheduling. Under a fixed software/hardware stack, and when stochastic operators consume BlazeFL-managed generators, this design is intended to support bitwise-identical repeated execution in high-concurrency settings. In practice, operators that internally rely on global RNG state must also be adapted to use framework-managed generators to achieve end-to-end determinism.

%% file: sec/3_system_design.tex
\section{System Design}
\label{sec:system_design}

BlazeFL targets a common but narrow setting: repeatable single-node FL simulation. Its design combines an execution model that reduces communication overhead with interface choices intended to keep experimental code easy to adapt. We describe the system through four aspects: shared-memory execution, controlled deterministic execution, protocol-based interfaces, and limited dependencies.

\subsection{Shared-Memory Execution via Free-Threading}

Most Python-based FL simulators achieve parallelism through multiple processes or external distributed runtimes. This is a practical response to the GIL, but it introduces process boundaries, runtime orchestration, and repeated parameter transfer across communication rounds. For single-node experiments, these costs can be substantial when communication and coordination dominate local computation.

BlazeFL primarily targets this setting by leveraging Python's free-threading support~\cite{pep703,pep779}. In the free-threaded mode, clients are executed as worker threads within a single process. The server prepares a downlink package containing the global state, and worker threads consume that package from shared memory rather than through cross-process serialization or an external object store. Client outputs are then returned to the server through the same shared address space. This design reduces communication overhead and keeps the execution model simple.

To separate the effect of the execution model from the effect of randomness control, BlazeFL also provides a process-based mode with shared-memory tensors. This allows us to compare free-threaded execution against a multiprocessing baseline that already removes most parameter-serialization cost.

\subsection{Controlled Deterministic Execution}

Parallel reproducibility in BlazeFL depends on controlling both client-local randomness and the order in which client results are materialized. Each client is associated with a dedicated RNG suite initialized from a deterministic seed schedule. This decouples client-local stochasticity---such as sampling, shuffling, augmentation, and dropout---from worker scheduling.

Determinism also depends on how client outputs are collected. In BlazeFL's default trainers, jobs are launched following the sampled client list, and the returned results are consumed through that same ordered job/future list rather than in completion order. The server therefore receives a stable buffer of client updates across repeated runs, avoiding one common source of floating-point divergence in parallel FL simulation.

Under a fixed software/hardware stack, and when stochastic operators consume BlazeFL-managed generators, this design supports bitwise-identical repeated execution in our high-concurrency setting. End-to-end determinism still requires user-defined components---for example, custom augmentations or server aggregation code---to avoid global RNG state and completion-order-dependent behavior.

\subsection{Protocol-Based Interfaces for Low-Coupling Experimentation}

Although BlazeFL is primarily a systems contribution, interface design was an important practical motivation. In many FL frameworks, researchers must adapt their training code to framework-specific base classes, lifecycle hooks, or runtime-owned object hierarchies. This can make small experimental changes unnecessarily invasive and can hinder reuse of existing PyTorch~\cite{pytorch} code.

BlazeFL therefore adopts protocol-based interfaces via Python's \texttt{typing.Protocol}~\cite{pep544}. Rather than requiring nominal inheritance, BlazeFL accepts any object as a valid server handler or client trainer as long as it implements the required methods. In practice, this means that ordinary training components can be integrated with minimal changes to class hierarchies or inheritance structure.

This choice does not directly improve throughput. Its role is instead to reduce framework lock-in and lower the cost of iterating on FL algorithms. We retain static type checking while keeping user code close to standard PyTorch training loops and data pipelines.

\subsection{Dependency Scope and Experimental Portability}

BlazeFL also keeps the runtime stack intentionally small. The core execution path relies on Python's standard libraries for \texttt{threading} and \texttt{multiprocessing} together with PyTorch~\cite{pytorch}, rather than depending on an external distributed scheduler, RPC stack, or object-store runtime. This smaller dependency surface simplifies setup and makes it easier to package, archive, and rerun experiments.

We view this as a practical aid to reproducibility rather than a formal guarantee. Minimal dependencies do not by themselves ensure identical results, but they reduce one common source of experimental fragility: changes in external runtime components that are orthogonal to the FL algorithm being studied.

%% file: sec/4_evaluation.tex
\section{Evaluation}
\label{sec:evaluation}

We evaluate BlazeFL along two axes: (1) wall-clock efficiency for single-node FL simulation and (2) deterministic repeatability under fixed experimental conditions. As a baseline, we use Flower~\cite{flower} with the Ray~\cite{ray} backend, a widely used open-source FL simulation framework. Because the two frameworks did not support the same Python/runtime stack in our environment, the reported results should be interpreted as a practical end-to-end comparison of supported configurations rather than as a same-interpreter microbenchmark.


\begin{figure*}[t]
    \centering
    \begin{subfigure}[b]{0.48\textwidth}
        \centering
        \includegraphics[width=\textwidth]{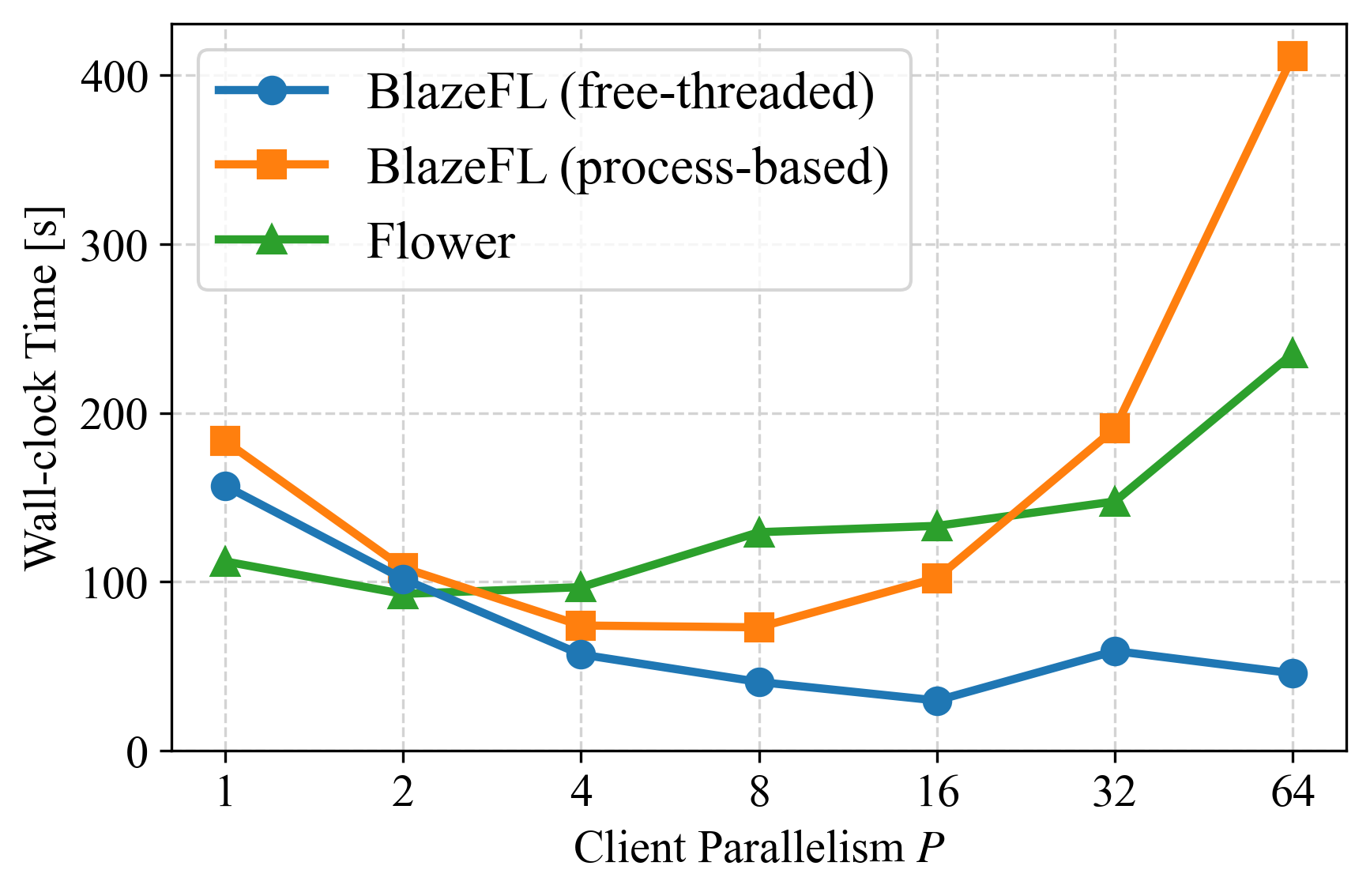}
        \caption{CNN}
    \end{subfigure}
    \hfill
    \begin{subfigure}[b]{0.48\textwidth}
        \centering
        \includegraphics[width=\textwidth]{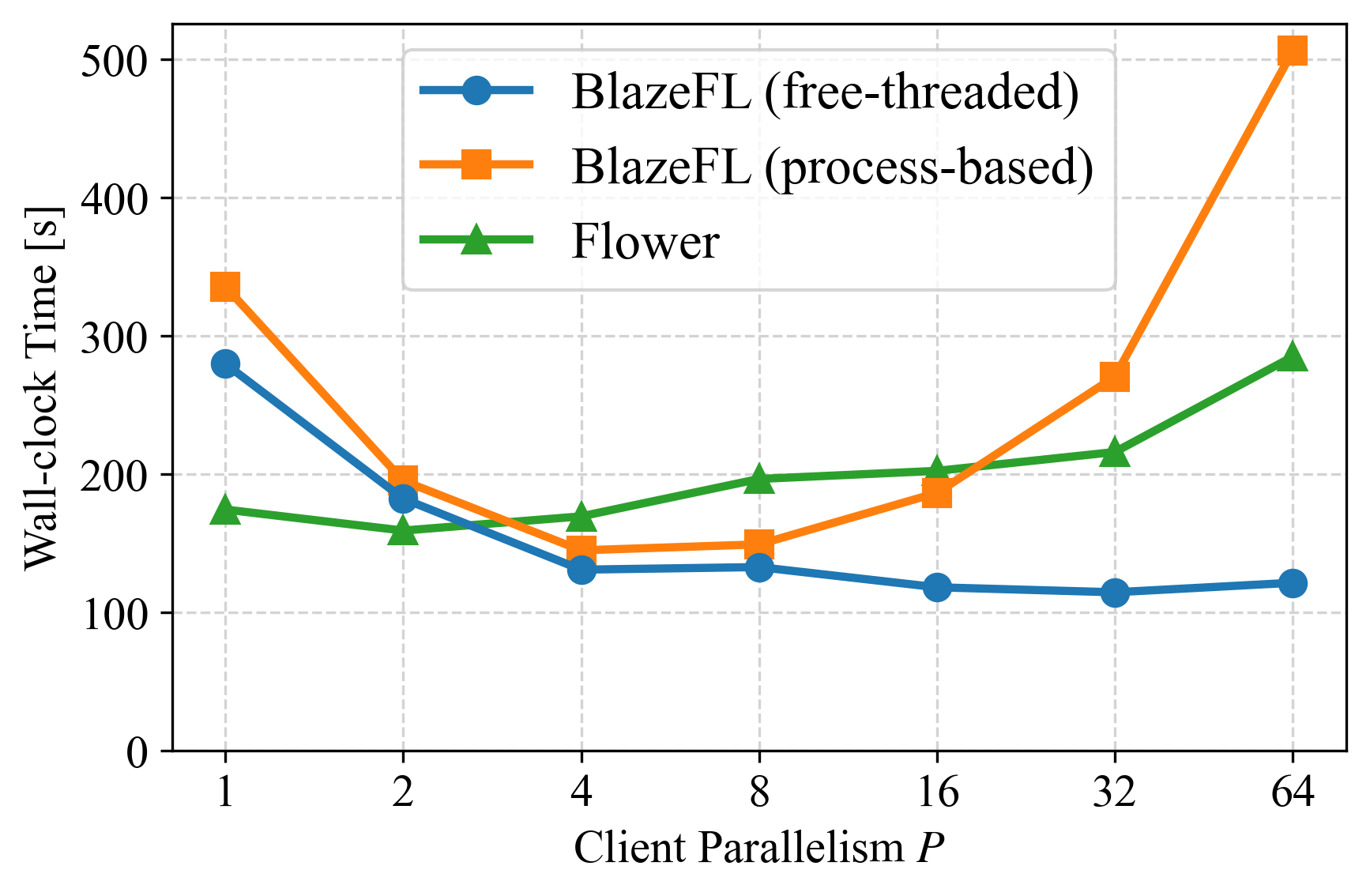}
        \caption{ResNet-18}
    \end{subfigure}
    
    \vspace{1em}

    \begin{subfigure}[b]{0.48\textwidth}
        \centering
        \includegraphics[width=\textwidth]{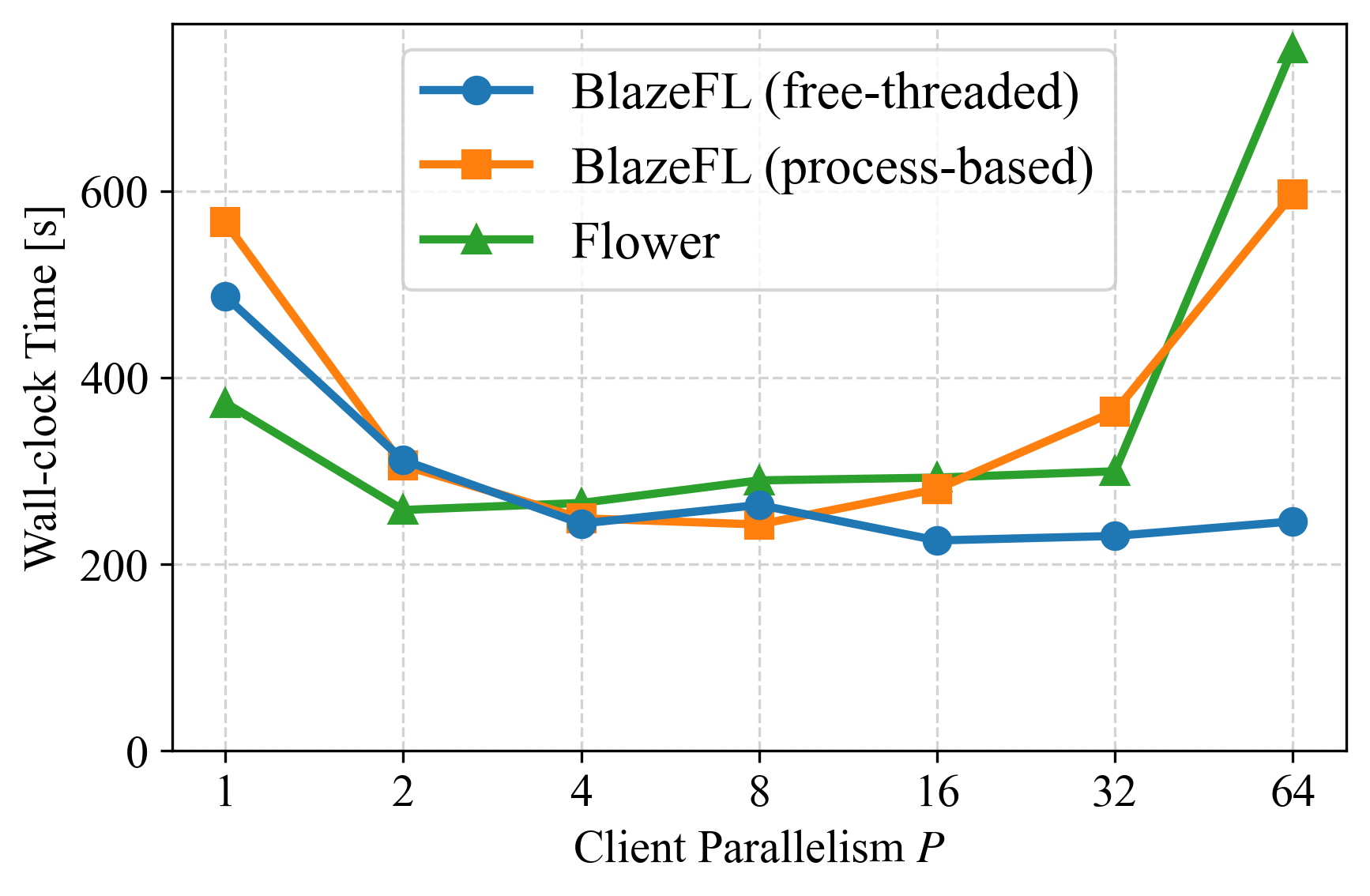}
        \caption{ResNet-50}
    \end{subfigure}
    \hfill
    \begin{subfigure}[b]{0.48\textwidth}
        \centering
        \includegraphics[width=\textwidth]{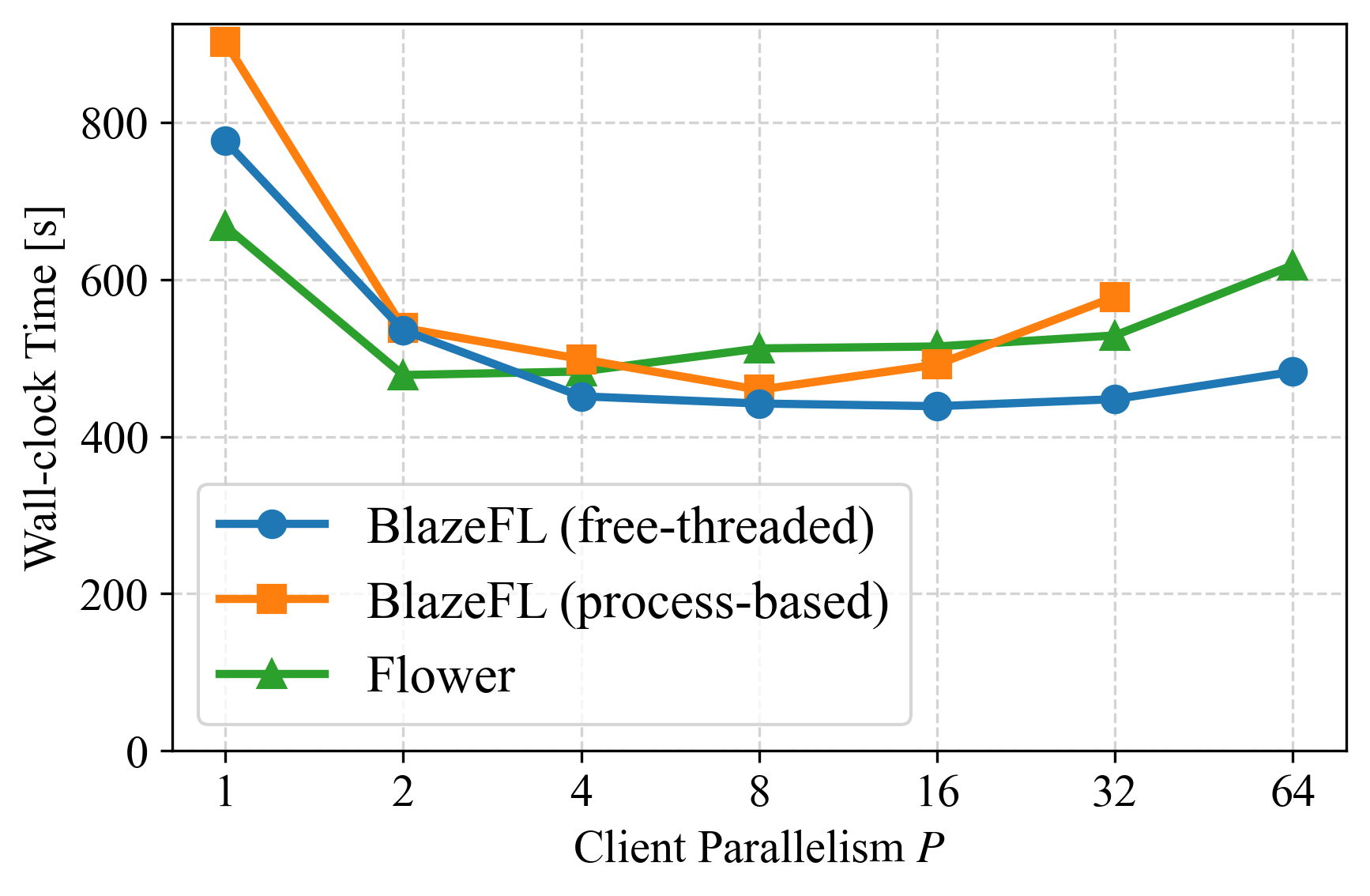}
        \caption{ResNet-101}
    \end{subfigure}
    
    \caption{Wall-clock time for five communication rounds on the high-performance server (48 CPU cores, NVIDIA H100) as a function of client parallelism $P$. Timings include client training, server aggregation, and global evaluation, but exclude dataset download and partition generation. Comparison of BlazeFL (free-threaded), BlazeFL (process-based shared memory), and Flower (Ray backend). Lower is better. Missing points for BlazeFL (process-based) indicate execution failure due to CUDA out-of-memory errors.}
    \label{fig:throughput_hpc}
\end{figure*}

\begin{figure*}[t]
    \centering
    \begin{subfigure}[b]{0.48\textwidth}
        \centering
        \includegraphics[width=\textwidth]{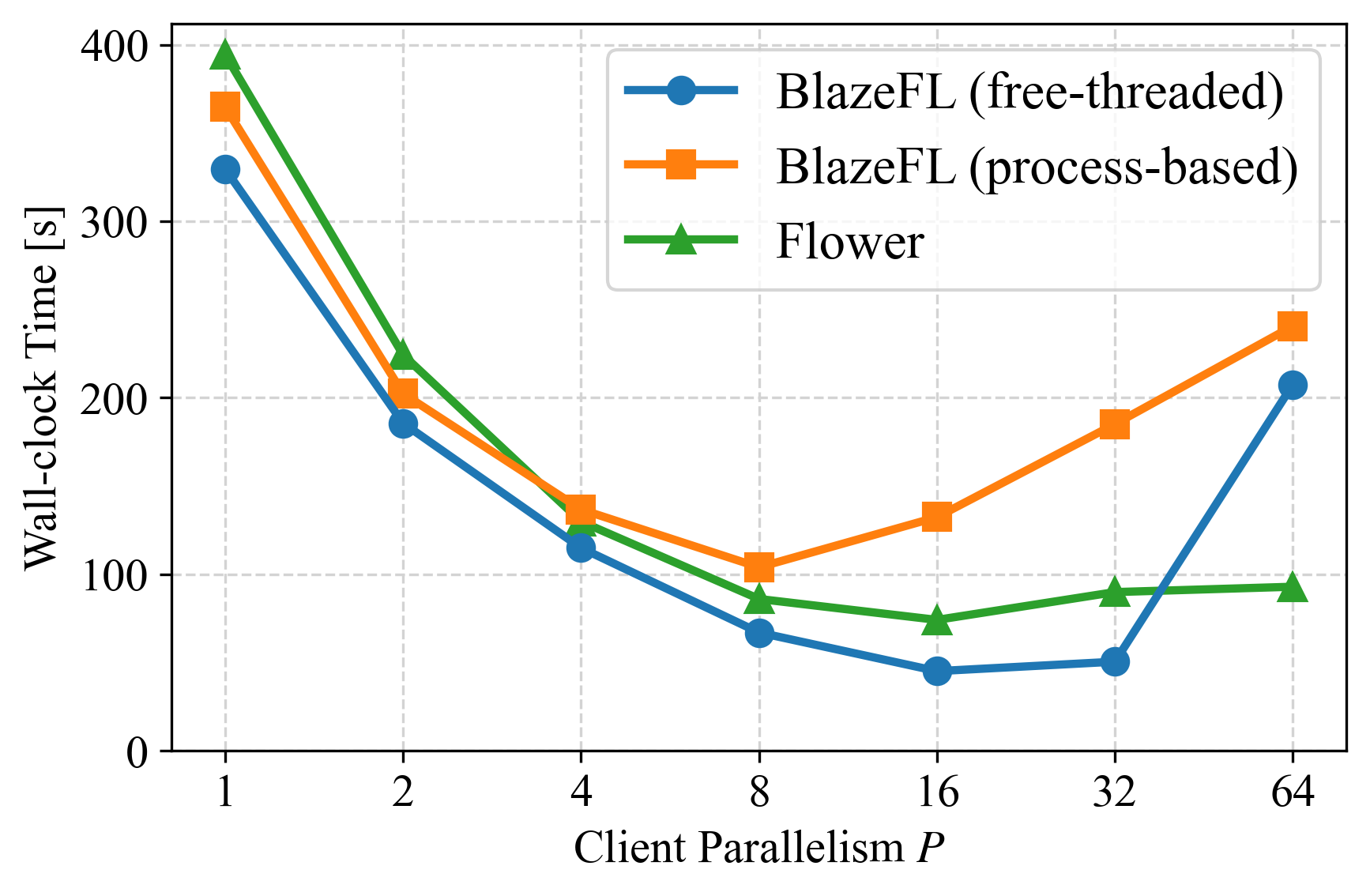}
        \caption{CNN}
    \end{subfigure}
    \hfill
    \begin{subfigure}[b]{0.48\textwidth}
        \centering
        \includegraphics[width=\textwidth]{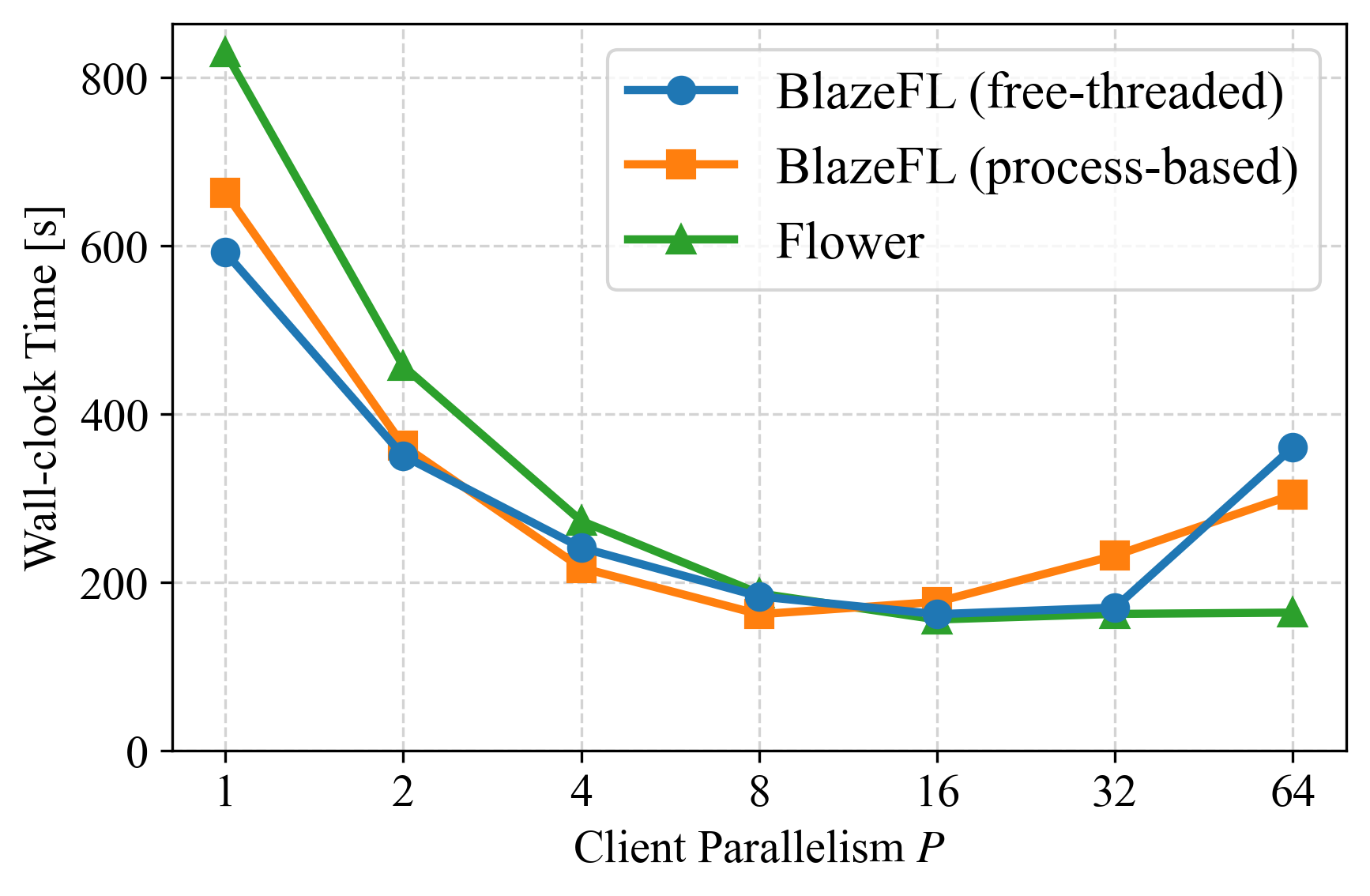}
        \caption{ResNet-18}
    \end{subfigure}
    
    \vspace{1em}

    \begin{subfigure}[b]{0.48\textwidth}
        \centering
        \includegraphics[width=\textwidth]{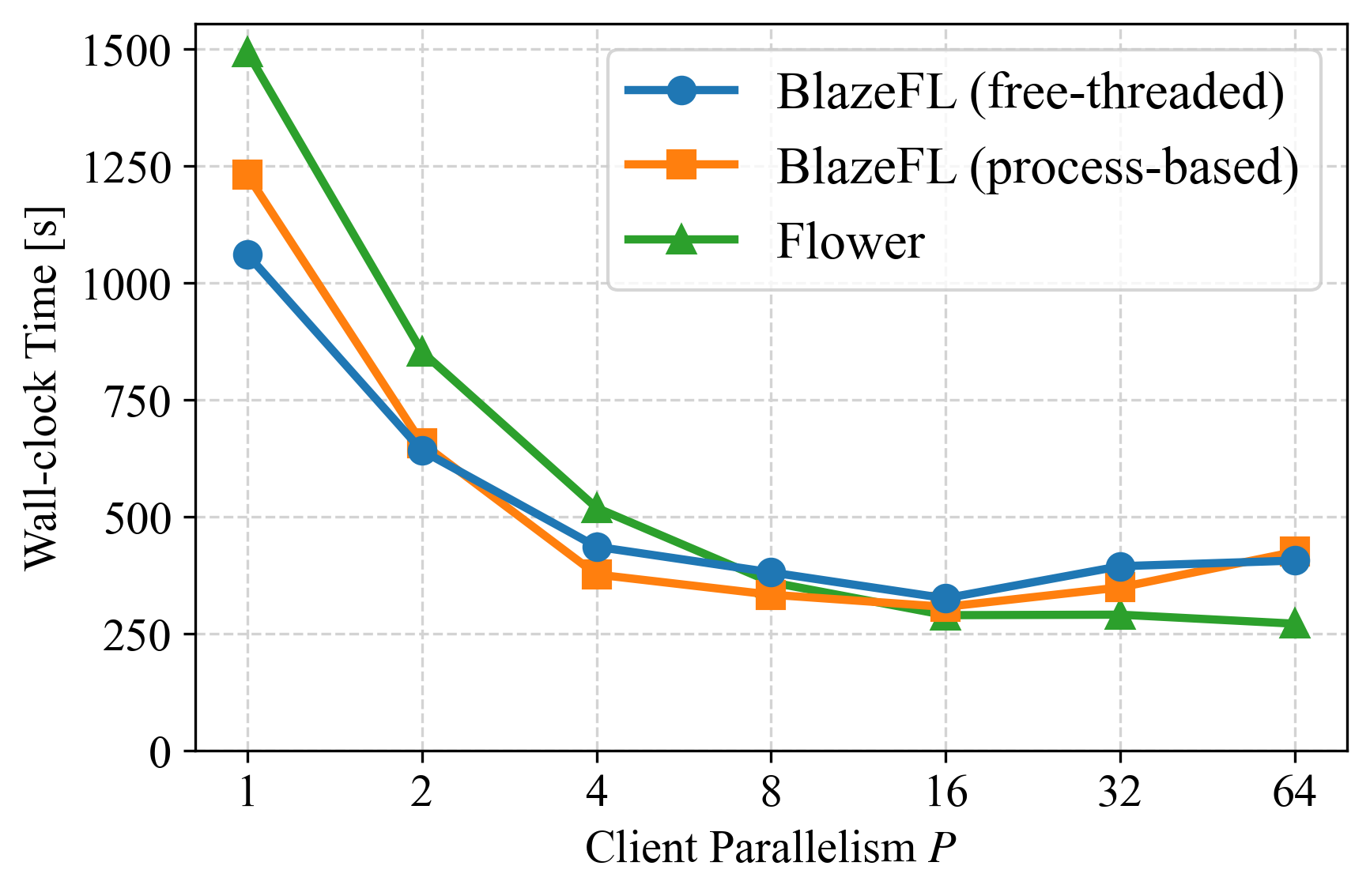}
        \caption{ResNet-50}
    \end{subfigure}
    \hfill
    \begin{subfigure}[b]{0.48\textwidth}
        \centering
        \includegraphics[width=\textwidth]{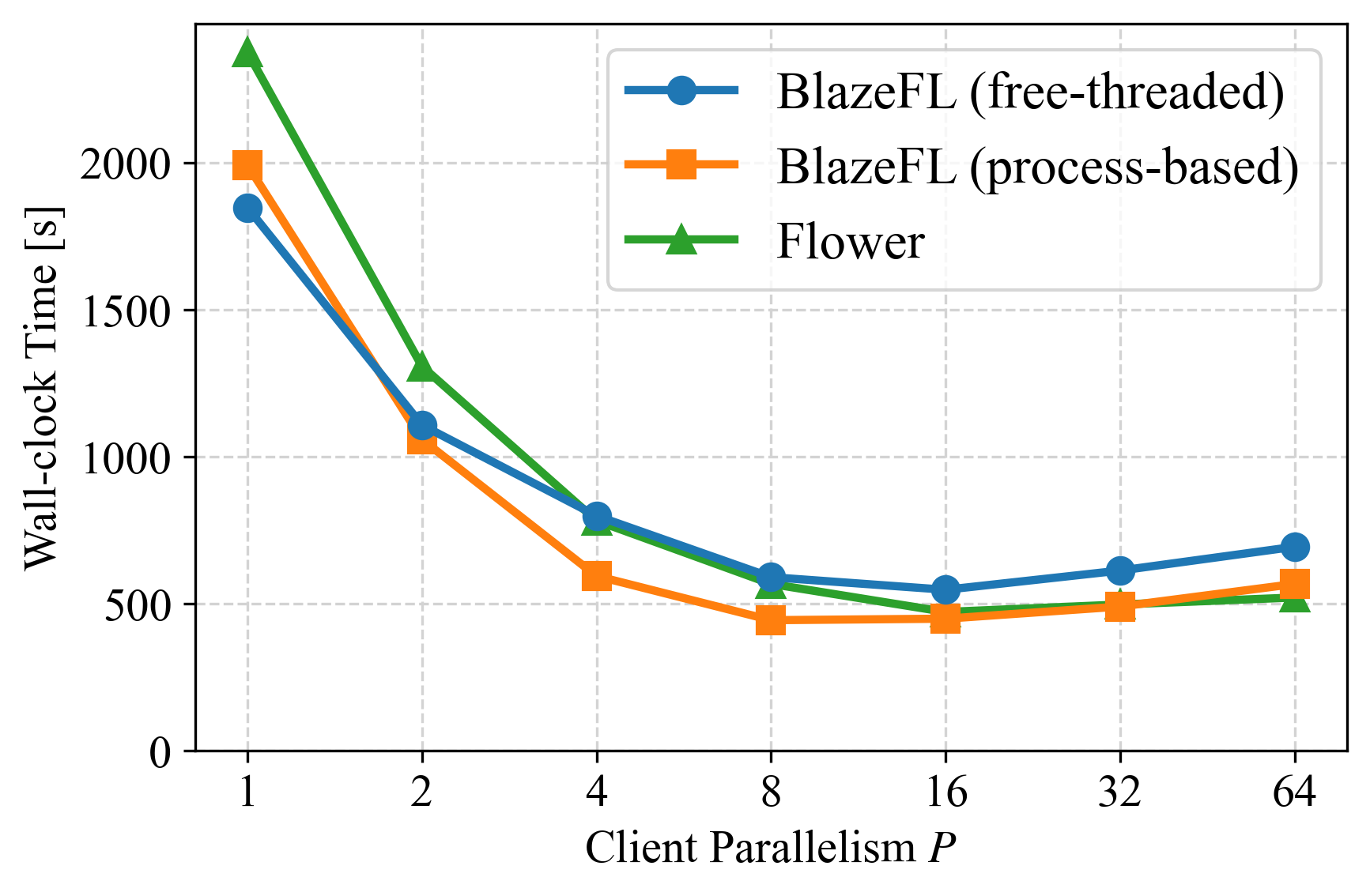}
        \caption{ResNet-101}
    \end{subfigure}
    
    \caption{Wall-clock time for five communication rounds on the workstation-class server (32 CPU cores, NVIDIA Quadro RTX 6000) as a function of client parallelism $P$. Timings include client training, server aggregation, and global evaluation, but exclude dataset download and partition generation. Comparison of BlazeFL (free-threaded), BlazeFL (process-based shared memory), and Flower (Ray backend). Lower is better.}
    \label{fig:throughput_workstation}
\end{figure*}

\subsection{Experimental Setup}
\label{subsec:experimental_setup}

Experiments were conducted on two representative hardware environments:

\begin{itemize}
    \item \textbf{High-performance server:} NVIDIA H100 GPU, 48 CPU cores, 192\,GB system memory.
    \item \textbf{Workstation-class server:} NVIDIA Quadro RTX 6000 GPU, 32 CPU cores, 256\,GB system memory.
\end{itemize}

BlazeFL was evaluated in a Python 3.14.3 environment, using the free-threaded build for the thread-based mode. Flower was evaluated in Python 3.13.7, which was the latest environment supported by its dependency stack in our setup. We revisit this limitation in \cref{subsec:current_ecosystem_maturity}. While the newer interpreter in Python 3.14 may provide marginal baseline speedups, the substantial performance gaps (e.g., up to 3.1$\times$) observed in communication-dominated workloads are primarily attributed to the elimination of IPC and serialization overheads, rather than interpreter-level optimizations.

We benchmarked CIFAR-10 image classification across 100 clients with a non-IID partition (two classes per client) and used FedAvg~\cite{fedavg} for server-side aggregation. In each experiment, the FL loop was executed for five communication rounds. Each selected client trained locally for 5 epochs on 500 samples, followed by server aggregation and evaluation on 10{,}000 test samples. We varied the degree of client parallelism as $P \in \{1, 2, 4, 8, 16, 32, 64\}$.

Timing measurements include only the five-round FL loop (client training, server aggregation, and global evaluation). Dataset download and partition generation are excluded from the reported wall-clock times.

We compared the following execution configurations:

\begin{itemize}
    \item \textbf{BlazeFL (free-threaded)}: Our proposed architecture, which uses Python's free-threading to execute clients as worker threads within a single process. Because all threads operate in a unified memory space, model parameters are accessed directly without serialization or inter-process communication.
    \item \textbf{BlazeFL (process-based)}: A multiprocessing implementation built with \texttt{torch.multiprocessing}, where model parameters are stored in shared-memory tensors to eliminate parameter serialization overhead.
    \item \textbf{Flower:} A standard process-based distributed execution using Ray.
\end{itemize}

To span communication-dominated and computation-dominated workloads, we evaluated a lightweight CNN and deeper residual networks including ResNet-18, ResNet-50, and ResNet-101~\cite{resnet}.

The throughput comparison uses the same high-level FL workload across frameworks, but not an identical saved partition file or identical dataset wrapper implementation, because Flower couples data handling to its own execution pipeline. Accordingly, the throughput results should be read as practical end-to-end measurements, whereas the reproducibility analysis below focuses on within-framework variation and hash agreement rather than absolute accuracy differences across frameworks.

\subsection{Throughput and Scalability}

We measured wall-clock time under varying degrees of client parallelism. \cref{fig:throughput_hpc} reports the results on the high-performance server, and \cref{fig:throughput_workstation} reports the results on the workstation-class server. In all figures, lower execution time indicates higher simulation throughput.

\subsubsection{High-performance server}

On the high-performance server (\cref{fig:throughput_hpc}), BlazeFL's free-threaded mode achieved the lowest execution times at moderate-to-high levels of parallelism. The largest gain appears for the lightweight CNN, where BlazeFL reached up to 3.1$\times$ lower wall-clock time than Flower. As the model size increases, the performance gap narrows but remains meaningful: the best observed speedups reach up to 1.4$\times$ for ResNet-18 and 1.1$\times$ for ResNet-50.

A consistent scaling trend is visible as $P$ increases. Flower tends to plateau and eventually degrade at higher worker counts, whereas BlazeFL's free-threaded mode continues to benefit from additional concurrency until hardware or framework-level limits are approached. The process-based BlazeFL implementation removes most parameter-serialization cost but remains slower than the free-threaded mode, suggesting that process management and cross-process coordination still contribute nontrivial overhead.

Overall, the results on this machine are consistent with BlazeFL's shared-memory execution being most beneficial in communication-dominated FL workloads.

\subsection{Workstation-class server}

On the workstation-class server (\cref{fig:throughput_workstation}), the qualitative trend is similar for lightweight workloads but less pronounced for larger models. BlazeFL remains clearly faster for the CNN model, indicating that reduced runtime coordination overhead is beneficial when local computation is modest. As model size increases, however, the gap narrows. At their best operating points, BlazeFL and Flower are comparable for ResNet-18, while Flower is slightly faster for ResNet-50 and ResNet-101. This behavior suggests that BlazeFL's advantage on this machine is concentrated in communication-dominated settings rather than compute-dominated ones.

While a deeper profiler-based analysis is left for future work, we hypothesize that this relative performance shift stems from PyTorch's internal C++ locks, particularly the global mutex within the CUDA caching allocator. On the high-performance server (80\,GB VRAM), abundant memory allows the allocator to operate on a fast path, keeping lock-holding times minimal. In contrast, the workstation's limited VRAM (24\,GB) under high concurrency likely forces the allocator into a slow path involving synchronous memory reclamation. When multiple threads submit compute-heavy kernels from a single process, this global allocator lock becomes a critical bottleneck.

Process-based execution (such as Flower or BlazeFL's process-based mode) bypasses this issue by assigning independent CUDA contexts to each worker, thereby avoiding single-process lock contention entirely. Therefore, in memory-intensive and VRAM-constrained scenarios under high concurrency, users may achieve better throughput by falling back to BlazeFL's process-based mode.

\subsection{Deterministic Behavior and Reproducibility}

\begin{table*}[t]
\centering
\caption{Repeated-run reproducibility at fixed parallelism ($P=32$) over 10 runs on the workstation-class server. Final accuracy standard deviation and round-wise SHA-256 hash agreement of the global model are reported. Because Flower and BlazeFL do not share identical data and partition pipelines, the table reports within-framework variability rather than absolute accuracy differences.}
\label{tab:reproducibility_runs}
\begin{tabular}{lcc}
\toprule
\textbf{Configuration} & \textbf{Final Acc. Std. Dev. [pp]} & \textbf{Round-wise Hash Agreement} \\
\midrule
Flower (no seed control) & 1.24 & No \\
Flower (global seed) & 0.18 & No \\
BlazeFL (process-based) & \textbf{0.00} & Yes \\
BlazeFL (free-threaded) & \textbf{0.00} & Yes \\
\bottomrule
\end{tabular}
\end{table*}

\begin{table}[t]
\centering
\caption{Reproducibility across degrees of client parallelism for BlazeFL (free-threaded). Using the same base seed, saved client partition, software stack, and five-round training schedule, we evaluate whether results remain identical as the number of parallel clients $P$ varies. Round-wise SHA-256 hashes match those of the $P=1$ reference run in all cases.}
\label{tab:reproducibility_parallelism}
\begin{tabular}{ccc}
\toprule
\textbf{$P$} & 
\makecell{\textbf{$\Delta$ Final Acc. [pp]}\\{(vs. $P=1$)}} &
\makecell{\textbf{Hash Agreement}\\{(vs. $P=1$)}} \\
\midrule
1  & --- & --- \\
2  & 0.0 & Yes \\
4  & 0.0 & Yes \\
8  & 0.0 & Yes \\
16 & 0.0 & Yes \\
32 & 0.0 & Yes \\
64 & 0.0 & Yes \\
\bottomrule
\end{tabular}
\end{table}

We next evaluate whether BlazeFL yields repeatable execution under fixed conditions. Unless otherwise noted, the results below are reported on the workstation-class server under a fixed software/hardware stack. We observed the same within-machine repeatability on the high-performance server, but we do not treat cross-machine bitwise identity as a target metric and therefore report one machine only.

\subsubsection{Repeated-run reproducibility at fixed parallelism}

We first performed 10 independent runs with $P=32$, a high-concurrency setting. We compared four configurations that differ in how randomness is controlled:
\begin{itemize}
    \item \textbf{Flower (no seed control):} default Flower execution without explicit seed control.
    \item \textbf{Flower (global seed):} Flower with manual initialization of \texttt{random}, \texttt{numpy}, and \texttt{torch} seeds during client setup.
    \item \textbf{BlazeFL (process-based):} BlazeFL with client-isolated RNG streams under multiprocessing.
    \item \textbf{BlazeFL (free-threaded):} BlazeFL with client-isolated RNG streams under free-threaded execution.
\end{itemize}

In BlazeFL, the same client-isolated RNG mechanism is used in both thread-based and process-based modes, and client results are consumed in a fixed sampled-client order. \cref{tab:reproducibility_runs} summarizes the resulting run-to-run variability.

Without seed control, Flower exhibited substantial variation, with a final-accuracy standard deviation of 1.24 percentage points. Manual global seeding reduced this variability to 0.18 percentage points, but did not eliminate it. In contrast, both BlazeFL configurations showed zero measurable final-accuracy variance across all 10 runs. The corresponding round-wise SHA-256 hashes of the global model were also identical across all runs for both BlazeFL modes, whereas Flower still exhibited hash mismatches. These results indicate that BlazeFL reproduces the entire training trajectory under a fixed software/hardware environment, not merely the final scalar accuracy.

\subsubsection{Reproducibility across degrees of parallelism}

We then directly tested whether BlazeFL's deterministic behavior changes with the degree of client parallelism. Using the same base seed, saved client partition, software stack, and training procedure, we ran the same five-round experiment with $P \in \{1, 2, 4, 8, 16, 32, 64\}$.

\cref{tab:reproducibility_parallelism} reports agreement with the $P=1$ reference run. All five round-wise SHA-256 hashes matched for every value of $P$, and the final test accuracy remained 20.53\% throughout. This result directly supports the claim that, within a fixed machine/software environment, BlazeFL's free-threaded execution is invariant to the degree of client parallelism in this benchmark.

\subsubsection{Diagnosing divergence in Flower}

\begin{figure}[t]
    \centering
    \includegraphics[width=\columnwidth]{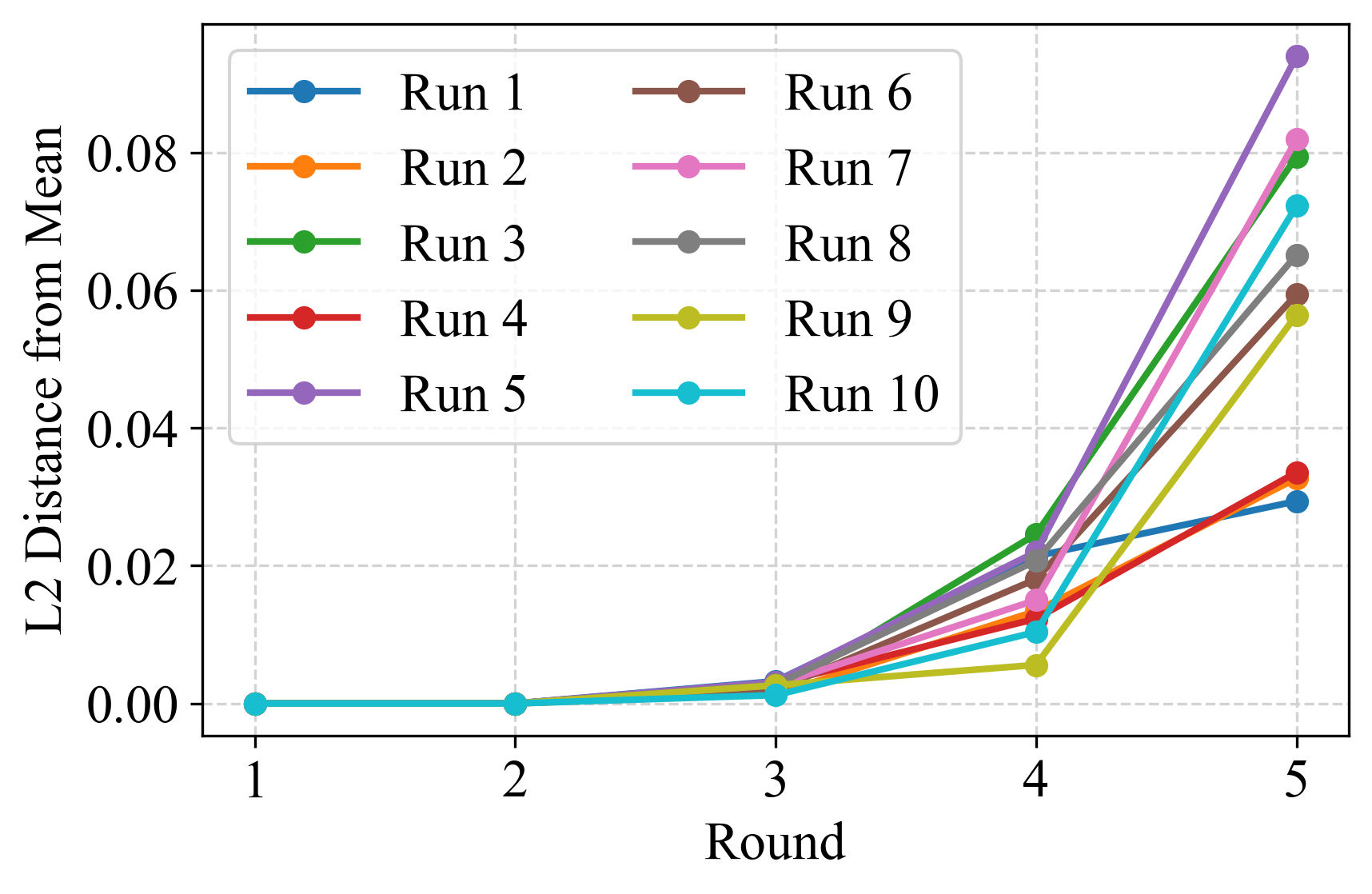}
    \caption{Accumulation of non-deterministic errors in Flower across 10 runs with manual global seeding. The $y$-axis represents the $L_2$ distance between each run's client logits and the mean logits of all runs at the start of each communication round. The trajectories fan out as floating-point rounding differences from non-deterministic aggregation order compound over time.}
    \label{fig:flower_divergence}
\end{figure}

To understand why Flower still diverges under manual global seeding, we tracked the logits produced for a specific data sample of a specific client across 10 independent runs. \cref{fig:flower_divergence} visualizes this divergence by plotting the $L_2$ distance of each run's output from the mean logits across all 10 runs at each communication round.

As shown in \cref{fig:flower_divergence}, the outputs are perfectly identical at the very beginning of training (Round 1). By Round 2, following the first server-side aggregation, microscopic differences on the order of $10^{-6}$ already emerge. While these initial discrepancies are too small to be visible on the macro-scale plot (appearing as zero), they act as the seed for divergence. Starting from Round 3, the trajectories begin to visibly fan out as these deviations are amplified by subsequent local training and aggregation phases, growing substantially in later rounds.

This behavior is consistent with completion-order-dependent accumulation. If client updates are materialized in worker-completion order rather than a fixed deterministic order, the sequence of floating-point additions in FedAvg can vary across runs due to slight differences in system scheduling and thread execution timings. Because floating-point addition is not strictly associative, these varied sequences produce slightly different rounding results during the aggregation step. As visualized by the fanning-out trajectories, these initially microscopic discrepancies compound over communication rounds, eventually leading to measurable differences in both model parameters and final accuracy.

%% file: sec/5_limitations.tex
\section{Limitations}

BlazeFL is designed for fast and repeatable \emph{single-node} FL simulation under controlled conditions. The results in \cref{sec:evaluation} should be interpreted within this scope.

\subsection{Single-Node Scope}

BlazeFL intentionally targets single-node simulation rather than general multi-node or production distributed training. Extending the framework to multi-node settings would introduce network communication, distributed synchronization, and additional runtime components, which would change both the performance model and the reproducibility model.

Accordingly, BlazeFL is best viewed as a tool for local prototyping, controlled benchmarking, and algorithmic debugging in FL research. It is not intended to replace general distributed FL runtimes.

\subsection{Determinism Depends on the Software/Hardware Stack}

Our reproducibility claims are limited to a fixed software/hardware environment. In our experiments, BlazeFL produced bitwise-identical repeated runs within the same machine and software stack, and also across degrees of client parallelism in the evaluated benchmark. However, we do not claim cross-machine or cross-platform bitwise identity.

In practice, differences in platforms, library versions, kernels, or hardware may change floating-point behavior or operator implementations even when the same seed is used. BlazeFL controls major framework-level sources of nondeterminism, but end-to-end reproducibility still depends on the surrounding numerical software stack.

\subsection{Generator Management in Vision Pipelines}

BlazeFL's deterministic execution relies on stochastic operations consuming framework-managed RNG streams. In computer-vision workloads, this requirement can be subtle because some preprocessing or augmentation operators may internally depend on global RNG state rather than an explicitly provided generator.

This issue is particularly relevant for transformations such as random crop or random flip. In such cases, per-client RNG isolation at the framework level is not by itself sufficient to guarantee end-to-end determinism under parallel execution. Users must ensure that vision-specific data pipelines are compatible with explicit generator management when strict reproducibility is required.

\subsection{Current Ecosystem Maturity}
\label{subsec:current_ecosystem_maturity}

BlazeFL benefits from Python's recent free-threading support, but the surrounding ecosystem is still maturing. Some third-party libraries, especially those with complex native extensions or tightly coupled distributed runtimes, may lag behind the latest free-threaded Python releases. This affects both usability and fairness of baseline comparisons, since not all FL frameworks can yet be evaluated under the same interpreter/runtime conditions.

We expect this limitation to weaken as ecosystem support improves. At present, however, BlazeFL should be understood as an early framework that is able to capitalize on free-threaded execution precisely because it keeps its runtime stack comparatively small.

%% file: sec/6_conclusion.tex
\section{Conclusion}
\label{sec:conclusion}

We presented BlazeFL, a lightweight framework for single-node federated learning simulation built around free-threaded shared-memory execution and controlled randomness management. BlazeFL reduces communication overhead by executing clients within a single process and exchanging model state through shared memory, while its client-isolated RNG design supports deterministic repeated execution under a fixed software/hardware stack.

Our experimental evaluation showed that BlazeFL can substantially reduce wall-clock time in communication-dominated workloads and that, in the evaluated benchmark, its execution remained bitwise-identical across repeated runs and across degrees of client parallelism on a single machine. These results suggest that BlazeFL provides a practical platform for fast and repeatable FL experimentation, especially in settings where local prototyping, benchmarking, and debugging are more important than general distributed deployment.

We hope BlazeFL serves as a useful systems tool for reproducible FL research and as an early example of how free-threaded Python can simplify high-concurrency machine learning simulation.

%% file: main.bib
@String(CVPR= {IEEE Conf. Comput. Vis. Pattern Recog.})

@String(CVPR  = {CVPR})

@misc{flower,
title={Flower: A Friendly Federated Learning Research Framework}, 
author={Daniel J. Beutel and Taner Topal and Akhil Mathur and Xinchi Qiu and Javier Fernandez-Marques and Yan Gao and Lorenzo Sani and Kwing Hei Li and Titouan Parcollet and Pedro Porto Buarque de Gusmão and Nicholas D. Lane},
year={2022},
eprint={2007.14390},
archivePrefix={arXiv},
primaryClass={cs.LG},
url={https://arxiv.org/abs/2007.14390}, 
}

@inproceedings {ray,
author={Philipp Moritz and Robert Nishihara and Stephanie Wang and Alexey Tumanov and Richard Liaw and Eric Liang and Melih Elibol and Zongheng Yang and William Paul and Michael I. Jordan and Ion Stoica},
title={Ray: A Distributed Framework for Emerging {AI} Applications},
booktitle={13th USENIX Symposium on Operating Systems Design and Implementation (OSDI 18)},
year={2018},
isbn={978-1-939133-08-3},
address={Carlsbad, CA},
pages={561--577},
url={https://www.usenix.org/conference/osdi18/presentation/moritz},
publisher={USENIX Association},
}

@manual{mpi,
author="{Message Passing Interface Forum}",
title="{MPI}: A Message-Passing Interface Standard Version 5.0",
url="https://www.mpi-forum.org/docs/mpi-5.0/mpi50-report.pdf",
year=2025,
month=JUN
}

@software{nccl,
author={{NVIDIA Corporation}},
title={{NCCL: NVIDIA Collective Communications Library}},
url={https://github.com/NVIDIA/nccl}
}

@misc{pfl-research,
title={pfl-research: Simulation Framework for Accelerating Research in Private Federated Learning},
author={Filip Granqvist and Congzheng Song and Aine Cahill and Rogier van Dalen and Martin Pelikan and Yi Sheng Chan and Xiaojun Feng and Natarajan Krishnaswami and Vojta Jina and Mona Chitnis},
year={2024},
URL={https://arxiv.org/abs/2404.06430}
}

@misc{horovod,
title={Horovod: fast and easy distributed deep learning in TensorFlow}, 
author={Alexander Sergeev and Mike Del Balso},
year={2018},
eprint={1802.05799},
archivePrefix={arXiv},
primaryClass={cs.LG},
url={https://arxiv.org/abs/1802.05799}, 
}

@inproceedings{pytorch,
author={Paszke, Adam and Gross, Sam and Massa, Francisco and Lerer, Adam and Bradbury, James and Chanan, Gregory and Killeen, Trevor and Lin, Zeming and Gimelshein, Natalia and Antiga, Luca and Desmaison, Alban and Kopf, Andreas and Yang, Edward and DeVito, Zachary and Raison, Martin and Tejani, Alykhan and Chilamkurthy, Sasank and Steiner, Benoit and Fang, Lu and Bai, Junjie and Chintala, Soumith},
booktitle = {Advances in Neural Information Processing Systems},
editor={H. Wallach and H. Larochelle and A. Beygelzimer and F. d\textquotesingle Alch\'{e}-Buc and E. Fox and R. Garnett},
pages={},
publisher={Curran Associates, Inc.},
title={PyTorch: An Imperative Style, High-Performance Deep Learning Library},
url={https://proceedings.neurips.cc/paper_files/paper/2019/file/bdbca288fee7f92f2bfa9f7012727740-Paper.pdf},
volume={32},
year={2019}
}

@misc{fedml,
title={FedML: A Research Library and Benchmark for Federated Machine Learning}, 
author={Chaoyang He and Songze Li and Jinhyun So and Xiao Zeng and Mi Zhang and Hongyi Wang and Xiaoyang Wang and Praneeth Vepakomma and Abhishek Singh and Hang Qiu and Xinghua Zhu and Jianzong Wang and Li Shen and Peilin Zhao and Yan Kang and Yang Liu and Ramesh Raskar and Qiang Yang and Murali Annavaram and Salman Avestimehr},
year={2020},
eprint={2007.13518},
archivePrefix={arXiv},
primaryClass={cs.LG},
url={https://arxiv.org/abs/2007.13518}, 
}

@misc{fedavg,
title={Communication-Efficient Learning of Deep Networks from Decentralized Data}, 
author={H. Brendan McMahan and Eider Moore and Daniel Ramage and Seth Hampson and Blaise Agüera y Arcas},
year={2023},
eprint={1602.05629},
archivePrefix={arXiv},
primaryClass={cs.LG},
url={https://arxiv.org/abs/1602.05629}, 
}

@InProceedings{resnet,
author={He, Kaiming and Zhang, Xiangyu and Ren, Shaoqing and Sun, Jian},
title={Deep Residual Learning for Image Recognition},
booktitle={Proceedings of the IEEE Conference on Computer Vision and Pattern Recognition (CVPR)},
month={June},
year={2016}
}

@techreport{pep544,
author={Ivan Levkivskyi and Jukka Lehtosalo and Łukasz Langa},
title={PEP 544 -- Protocols: Structural subtyping (static duck typing)},
year={2017},
type={Python Enhancement Proposals},
number={544},
institution={Python Software Foundation},
url={https://peps.python.org/pep-0544/}
}

@techreport{pep703,
  author      = {Sam Gross},
  title       = {{PEP 703 -- Making the Global Interpreter Lock Optional in CPython}},
  year        = {2023},
  type        = {Python Enhancement Proposals},
  number      = {703},
  institution = {Python Software Foundation},
  url         = {https://peps.python.org/pep-0703/}
}

@techreport{pep779,
  author      = {Thomas Wouters and Matt Page and Sam Gross},
  title       = {PEP 779 -- Criteria for supported status for free-threaded Python},
  year        = {2025},
  type        = {Python Enhancement Proposals},
  number      = {779},
  institution = {Python Software Foundation},
  url         = {https://peps.python.org/pep-0779/}
}
